%% file: cvpr.tex
\documentclass[final]{cvpr}

\usepackage{times}
\usepackage{epsfig}
\usepackage{graphicx}
\usepackage{amsmath}
\usepackage{amssymb}

\usepackage[pagebackref=true,breaklinks=true,colorlinks=false,bookmarks=false]{hyperref}
\def\cvprPaperID{21} 
\def\confYear{FGVC8 2021}

\begin{document}

\title{United We Learn Better: \\ Harvesting Learning Improvements From Class Hierarchies Across Tasks}

\author{Sindi Shkodrani \\
\and
Yu Wang\\
\and
Marco Manfredi \\

\and
Nóra Baka \\

\and
TomTom\\
Amsterdam, The Netherlands\\
{\tt\small [firstname.lastname]@tomtom.com}

}

\maketitle

\begin{abstract}
   Attempts of learning from hierarchical taxonomies in computer vision have been mostly focusing on image classification. Though ways of best harvesting learning improvements from hierarchies in classification are far from being solved, there is a need to target these problems in other vision tasks such as object detection. As progress on the classification side is often dependent on hierarchical cross-entropy losses, novel detection architectures using sigmoid as an output function instead of softmax cannot easily apply these advances, requiring novel methods in detection. In this work we establish a theoretical framework based on probability and set theory for extracting parent predictions and a hierarchical loss that can be used across tasks, showing results across classification and detection benchmarks and opening up the possibility of hierarchical learning for sigmoid-based detection architectures. \footnote {Code is available at {\tt\scriptsize \url{https://github.com/sindish/unitedwelearn}}.}    
\end{abstract}

\section{Introduction}

Learning from hierarchical taxonomies of label semantics in computer vision is not a recent problem, yet most existing methods have been tackling image classification only \cite{silla2011survey}. As advanced vision applications often make use of more complex tasks such as object detection or semantic segmentation, there is an open need to further study hierarchical learning beyond image classification.

Taxonomic information contained in hierarchies can add semantics and act as extra supervision in deep networks \cite{brust2019integrating, lee2018hierarchical}. In addition, research has shown that hierarchical groupings can be implicitly learned by deep features without any specific hierarchical labels provided \cite{sivic2008unsupervised, bilal2017convolutional, akata2015evaluation}, indicating that explicitly enforcing hierarchical learning might benefit deep features. 


Another compelling reason to leverage these taxonomies is the need to minimize the severity of mistakes. Misclassifying a \textit{pedestrian} as a \textit{cyclist} is not as big of a mistake as misclassifying it as a \textit{tree}, particularly in safety-critical applications such as autonomous driving. Here, hierarchical evaluation measures are pivotal \cite{making_better_mistakes, wu2016learning, verma2012learning}. Semantic hierarchies have been used for automatic image annotation \cite{bannour2012hierarchical, fan2007hierarchical}. In applications automated by machine learning models and corrected by humans, a small distance mistake might be less costly to correct than a large distance mistake, reducing the man-hour cost of annotation.



In this work we enable hierarchical learning for object detection, where sigmoid-based focal loss is predominantly used ~\cite{retinanet}, as current methods mostly tackle softmax-based cross-entropy losses. Our proposed framework is built on a set theory-based interpretation of hierarchical probabilities, which not only enables sigmoid-based detection methods, but also explains softmax as a special case, yielding a simpler formulation of a hierarchical softmax cross-entropy that achieves superior results to previous approaches. 

\textbf {Hierarchical classification} is a well studied problem, with most approaches enforcing learning of hierarchies by either changing the network structure ~\cite{bcnn, hdcnn, bilal2017convolutional, goo2016taxonomy} or solely the loss function ~\cite{goyal2020hierarchical, brust2019integrating, garnot2020leveraging, dhall2020hierarchical, mettes2019hyperspherical,making_better_mistakes, barz2019hierarchy, hierlossproblems}. Architecture-based approaches are hard to generalize to the detection scenario, as object detection networks generally consist of a heavy-weight general feature extractor backbone, FPN~\cite{lin2017feature}, whereas loss-based approaches from classification are not easily applicable to detection, as there we have to handle multiple predictions for a single object. 

\textbf{Hierarchical object detection} was introduced by YOLOv2~\cite{YOLO9000}, where only the classification head is fine-tuned with a large number of objects and relies heavily on softmax, unlike recent popular detection methods, such as the RetinaNet \cite{retinanet} that rely on sigmoid. ~\cite{bu2019learning} enable multi-label classification by the detector, with labels being sometimes synonyms, but can also be used in a hierarchical setting. Focal loss \cite{retinanet} did not work in their case. Recently, a Wasserstein loss was proposed for weighting detector errors by severity~\cite{wasserstein_obj_detection}, similarly to the Earth Mover's Distance used by~\cite{hou2016squared} and only applied to a softmax-based Faster-RCNN ~\cite{ren2015faster}. We show that hierarchical training can indeed be done using focal loss. 

\textbf{Error severity evaluation}
In classification the error severity can be measured by e.g. the lowest common ancestor of the ground truth label and the predicted label in the hierarchical tree~\cite{making_better_mistakes, kosmopoulos2015evaluation}. In detection, mAP ~\cite{mscoco} is the standard metric, combining localization and classification accuracy. For evaluation with hierarchies, ~\cite{bu2019learning} simply duplicates each box for every hierarchy level with corresponding level label, yielding an mAP score with this enhanced ground truth. We propose to calculate mAP at every level, and thereby enable deeper analysis of the performance.

Summarized, our contributions are threefold. First, we propose a theoretical framework of hierarchical losses that enables hierarchical learning for modern object detectors and demonstrate improvements over existing benchmarks. Next, we demonstrate the validity of this interpretation for hierarchical image classification, proposing a simpler hierarchical cross-entropy formulation and improving over existing hierarchical baselines. Last, we introduce a hierarchical evaluation method for object detection which enables validating learning improvements in the higher levels of hierarchy.

\section{Method}\label{section:method}
\subsection{Softmax hierarchy aggregation}

Classification networks since the AlexNet~\cite{krizhevsky2017imagenet} often use softmax to map network output to respective predicted probabilities. It is quite for classification, as it normalizes the output of a network to a probability distribution over predicted output classes. As it's a distribution and sums up to one, this makes it suitable to use in loss functions such as cross-entropy to optimize the predictions into matching one-hot encoded classification labels. As an output function, softmax enforces a hard assignment to classes, where each instance can only belong to one. 

For class hierarchies this means that to extract the parent probabilities from the probability distribution of the children, we can simply sum the probabilities of the respective children. The parent probabilities will still add up to 1 and be a probability distribution. Thus for many disjoint children classes occurring, the probability of a parent class would be

\begin{equation}
    p(C_i^{l+1}) = \sum _{k=1}^{K} p(C_{k}^{l})
        \label{eqn:sum}
\end{equation}

where $C_i$ is the parent class and $C_{k}, k \in {0, ... K}$ are its respective children classes of the lower level. The network only needs to predict leaf-level probabilities. The higher levels can be computed with equation \ref{eqn:sum} knowing leaf-level $p(C_{k}^{1})$  from network predictions.

\begin{figure*}
\begin{center}
\includegraphics[width=\linewidth]
                   {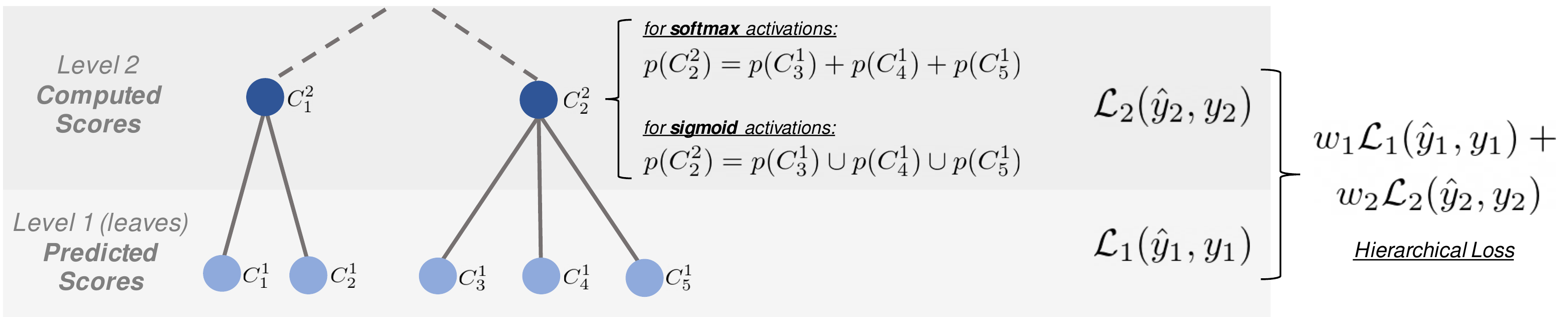}

\end{center}
    \caption{Overview of our method for a simple two-level hierarchy. Starting from the network output at the leaf level, we compute aggregated scores for parent nodes with either a sum (in the softmax case) or with probability union (sigmoid case). The proposed hierarchical loss is obtained as the weighted sum of the per-level losses.}
\label{fig:probabilities}
\end{figure*}

\subsection{Sigmoid hierarchy aggregation}

Sigmoid is a function defined for all real input values and is bounded between 0 an 1. This makes it suitable for extracting probabilities from network output logits, though unlike softmax, it does not normalize them to a probability distribution. This makes sigmoid suitable for multi-class outputs \cite{kuznetsova2020open}. RetinaNet \cite{retinanet} is a notable example of a modern detection architecture showing that using sigmoid allows for greater numerical stability due to relaxing the single class assumption. RetinaNet derivatives and many other detection methods \cite{zhou2019objects, law2018cornernet} have used sigmoid from then on. 

Sigmoid, unlike softmax, allows multi-class predictions and the output is not a probability distribution, thus classes are not assumed mutually exclusive. If we were to add the children probabilities, we'd get values higher than 1. To accommodate the sigmoid case, we interpret class probabilities in light of probability and set theory. Given a set of children classes for each parent, we can say that a parent class occurs if either of the children classes occur. From set theory, this is the union of probabilities of children classes occurring:

\begin{equation}
    p(C_i^{l+1}) = \bigcup _{k=1}^{K} p(C_{k}^{l})
    \label{eqn:union}
\end{equation}

This is in line with the softmax reasoning. With softmax the predicted probabilities assign each instance to a single class, assuming the sets disjoint. Two events are mutually exclusive or disjoint if they cannot occur at the same time. Since for two disjoint events $ A \cap B = \emptyset $, the union probability rule $ P(A \cup B) = P(A) +  P(B) + P(A \cap B)$ becomes $ P(A \cup B) = P(A) +  P(B)$. This leads to equation \ref{eqn:sum}, making the softmax sum a special case of the union. 

In order to extract parent probabilities from sigmoid output children probabilities, we have to compute the probability unions. Probability unions can be computed with the inclusion-exclusion principle: 

\begin{equation}
\small
\begin{split}
 \left| \bigcup_{1 \leq i \leq n}  A_i\right|
 = & \sum_{1 \leq {i_1} \leq n} \left| A_{i_1}\right| 
 - \sum_{1 \leq {i_1} \leq {i_2} \leq n} \left| A_{i_1} \cap A_{i_2}\right| \\
&  + \sum_{1 \leq {i_1} \leq {i_2} \leq{i_3} \leq n}
\left| A_{i_1} \cap A_{i_2}\cap A_{i_3}\right|   - \ldots \\
&  + (-1)^{n+1} \left| \bigcap_{i=1}^n A_i \right|
\end{split}
\end{equation}

We apply the inclusion-exclusion principle to compute parent probabilities from sigmoid probability outputs of children. 

\begin{equation}
\small
\begin{split}
    \bigcup _{k=1}^{K} p(C_{k}^{l}) = & \sum_{1 \leq {k_1} \leq K} p(C_{k_1}^{l})  \\
                                &    - \sum_{1 \leq {k_1} \leq {k_2} \leq n} p(C_{k_1}^{l})  p(C_{k_2}^{l})   \\
                                 &   + \sum_{1 \leq {k_1} \leq {k_2} \leq{k_3} \leq n}  p(C_{k_1}^{l})  p(C_{k_2}^{l})     p(C_{k_3}^{l}) \\
                                &    - \ldots + (-1)^{k+1} \sum_{1 \leq {k_1} ... {k_K}\leq n} \prod_{j\in K} p(C_{k_j}^{l}) 
\end{split}
\end{equation}

This computation has exponential complexity, making it costly for hierarchies with large number of children per parent. We implement these with matrix operations, preserving fast and differentiable computations.

\subsection{Multi-level hierarchical loss functions}

With aggregated predictions as explained above, we compute the multi-level hierarchical loss as:

\begin{equation}
    \mathcal{L}_{hierarchical} = \sum_{l=1}^L w_l \mathcal{L}_l (\hat{y}_l, y_l) 
\end{equation}

where  $w_l$ is a level weight, $\mathcal{L}_l$ is the respective level loss and $\hat{y}_l$, $y_l$ the respective layer predictions aggregated from the children and the layer ground truth coming from the tree mapping of the ground truth leaf class. We illustrate our loss in Figure \ref{fig:probabilities}. Using the principles above, we implement a multi-level hierarchical cross-entropy and a multi-level hierarchical focal loss on existing classification and detection architectures.

\subsection{Hierarchical evaluation for object detection}
To evaluate hierarchical improvements in our detection experiments we introduce a multi-level mean average precision. This is simply the mAP computed at every level of the hierarchy, with aggregated predictions and converted ground truth. Since the predicted bounding boxes for evaluation are selected by non-maximum suppression, if we were to only change the class labels of the boxes, we would disregard the effect of the aggregation on boxes discarded by the NMS. Because of this, we re-compute NMS at every level of the hierarchy, enabling a corrected set of boxes to emerge after higher level score aggregation. Our multi-level NMS enables wrong leaf-level predictions to be corrected if the joint probability of siblings leads to a more accurate parent class.


\input{experiments.tex}

\section{Conclusion}

In this work we introduced a hierarchical learning method for object detection by generalizing over the classification case. We suggested a novel way to aggregate network predictions up the hierarchical tree that yields a simpler hierarchical cross-entropy loss as well as the first attempt at a hierarchical focal loss. Our proposed multi-level mean average precision allows us to evaluate the improvement of mistake severity in detection predictions. We evaluated improvements with our hierarchical losses across detection and classification benchmarks.

{\small
\bibliographystyle{ieee_fullname}
\bibliography{egbib}
}

\clearpage

\section*{Appendix}

\subsection*{A: Implementing Differentiable Aggregations}
To implement the hierarchical cross entropy loss, we aggregate leaf-level predicted probabilities into parent-level classes. For softmax-based losses, we aggregate by adding children probabilities while for sigmoid-based ones by using probability unions. 

\subsubsection*{Hierarchical Sum Aggregations}

To aggregate by sum, we need to convert $p(C_{k}^{l})$ probabilities of level $l$ into level $l+1$ probabilities $p(C_i^{l+1})$. $k \in 0, 1, ... K-1$ are the children classes and $i \in {0, 1, ... I-1}$ are the parent classes, where $K$ is the number of children classes, $I$ the number of parent classes and $K \leq I$. To implement the sum aggregation we construct a $K \times I$ one-hot matrix, where for each row we set to 1 the column that corresponds to the parent index. 

In this way, we can simply apply matrix multiplication between class probabilities in the form $\textit{batch size} \times K$ to the $K \times I$ conversion matrix and get $\textit{batch size} \times I$ parent level probabilities.

\subsubsection*{Implementing Hierarchical Union Aggregations}

To aggregate by union, we resort to the inclusion-exclusion principle of probability unions, where we step by step should add all children probabilities, subtract multiplied pairs, add multiplied triplets, subtract quadruples and so on.   

To keep operations differentiable, from the original $K \times I$ level transition matrices mentioned in the sum case above, we generate new matrices containing all combinations of children for each parent. So instead of one  $K \times I$  one-hot matrix, for each parent we now multiply the original probabilities to many matrices corresponding to the combinations possible at each addition or subtractions step. E.g: at the pair subtraction step, we make $n$ matrices corresponding to the number of 2-children combinations in a parent with $N$ children classes; at triplet addition step we make $m$ matrices corresponding to the number of 3-children combinations from $N$ and so on. The time complexity is $\mathcal{O}(2^N \times N)$. This is not cheap, however considering that $N \leq K \leq I$ and that the number of classes keeps decreasing while going higher in the hierarchy tree, in practice the effect on training time is minimal.

\subsection*{B: Comparing to other hierarchical cross-entropy losses}

Here, we compare our hierarchical cross-entropy formulation numerically to the formulation in \cite{making_better_mistakes}. The authors formulate the node probabilities as:

\begin{equation*}
\begin{split}
    p(C) &= \prod_{h=0}^{H-1} p(C^{(h)}|C^{(h+1)}) \\
         &= \prod_{h=0}^{H-1} \frac{\sum_{A \in \text{Leaves}(C^{(h)})} p(A)}{\sum_{B \in \text{Leaves}(C^{(h+1)})} p(B)}
\end{split}
\end{equation*}

where H is the depth of node C. They use these conditional probabilities for the hierarchical loss computation at every level. Thus for every parent level, the node probability is conditional to the children probabilities. So at every level of the loss computation, the $p(C)$ values couple the current node level to the next level leaves.

In contrast, we simply use the sum-aggregated probabilities at the respective level without interpreting them as conditional.

\begin{equation*}
    p(C_i^{l+1}) = \sum _{k=1}^{K} p(C_{k}^{l})
        \label{eqn:sum_app}
\end{equation*}

where $k \in {0, ... K}$ are the children classes of the next level. We think this simpler interpretation generates cleaner gradients from the loss, enabling the preservation of Top-1 accuracy while improving hierarchical distance metrics in the results.


\subsection*{C: Discussion about datasets}

As discussed in the paper, our hierarchical loss consistently improves classification errors in object detection on COCO \cite{mscoco}. However, we observe in our visual results that the overall number of classification errors in COCO is not so high, as most errors seem to come from difficult, overlapping instances rather than misclassification. We had to resort to COCO due to the availability of WordNet\cite{wordnet} hierarchies. For future work, using detection datasets with more fine-grained classes is recommended for showing larger hierarchical improvements, while there is still an open need for datasets targeted at hierarchical object detection.

\end{document}

%% file: experiments.tex
\section{Experiments}




\subsection{Hierarchical focal loss for object detection}

\begin{table*}
\begin{center}
\resizebox{\textwidth}{!}{
\begin{tabular}{|l|c|c|c|c|c|c|c|c|}
\hline
Method & mAP (leaf) & mAP (level 2) & mAP (level 3) & mAP (level 4) & mAP (level 5) & mAP (level 6) & mAP (level 7) & mAP (level 8)\\
\hline\hline
RetinaNet (ResNet50-FPN 1x) & 37.30 & 37.37 & 36.84  & 36.57 & 37.42 & 40.44 & 44.51 & 46.06 \\
Ours (ResNet50-FPN 1x ) & \textbf{37.44}  & \textbf{37.49}  & \textbf{37.01} & \textbf{36.96}   & \textbf{37.70} &  \textbf{40.90}  & \textbf{44.89}  & \textbf{46.43}  \\
\hline
RetinaNet (ResNet50-FPN 3x) & 38.60  & \textbf{38.54} & 38.18 & 38.10 & 39.15 & 42.08 & 45.56 & 47.26  \\
Ours (ResNet50-FPN 3x ) & \textbf{38.63}  &  38.51  & \textbf{38.19}   & \textbf{38.22}  & \textbf{39.17}  & \textbf{42.27}  & \textbf{45.78}  & \textbf{47.45}  \\
\hline
RetinaNet (ResNet101-FPN)  & 40.24  & 40.20  & 39.83  & 39.83  & 40.86  & 43.61 & 47.29  & 49.05 \\
Ours (ResNet101-FPN)  & \textbf{40.28}  & \textbf{40.25}  & \textbf{40.03}  & \textbf{40.23}  & \textbf{41.40} & \textbf{44.06} & \textbf{47.59} & \textbf{49.35} \\
\hline
\end{tabular}
}
\end{center}
\caption{Results on COCO object detection benchmark with our hierarchical focal loss (1x and 3x indicate shorter and longer training schedules from standard object detector training in Detectron2 ~\cite{detectron2}).}
\label{tab:coco}
\end{table*}

\begin{table*}
\begin{center}
\resizebox{\textwidth}{!}{
\begin{tabular}{|l|c|c|c|c|c|}
\hline
Method & Hier. dist. mistake & Avg. hier. dist. @1 & Avg. hier. dist. @5 & Avg. hier. dist. @20 & Top-1 error \\
\hline\hline

CROSS-ENTROPY & 2.49 ± 0.0051 & 1.16 ± 0.0031 & 1.98 ± 0.0027 & 2.92 ± 0.0039 & 46.60 ± 0.1472 \\
Bertinetto et al \cite{making_better_mistakes} HXE ($\alpha=0.1$)  & 2.43 ± 0.0050 & 1.15 ± 0.0067 & 1.87 ± 0.0059 & 2.74 ± 0.0058 & 47.27 ± 0.2095 \\
Bertinetto et al \cite{making_better_mistakes} HXE ($\alpha=0.5$) & 2.27 ± 0.0059 & 1.25 ± 0.0077 & 1.71 ± 0.0055 & \textbf{2.38 ± 0.0043} & 55.07 ± 0.2204 \\
\hline
Our HXE (leaf-focused weighting) &  2.44 ± 0.0014 & \textbf{1.09 ± 0.0021} & 1.90 ± 0.0025 & 2.89 ± 0.0031 & \textbf{44.55 ± 0.0701} \\
Our HXE (hier-focused weighting) & \textbf{2.27 ± 0.0048} & 1.11 ± 0.0042 & \textbf{1.61 ± 0.0038} & 2.39 ± 0.0040 & 49.14 ± 0.0961 \\
\hline
\end{tabular}
}
\end{center}
\caption{Results on iNaturalist classification benchmark.}
\label{tab:inaturalist}
\end{table*}

\begin{table}
\begin{center}
\resizebox{\linewidth}{!}{
\begin{tabular}{|l|c|c|c|}
\hline
Method & mAP (leaf) & mAP (level 2) & mAP (level 3) \\
\hline\hline
Faster-RCNN (ResNet50-FPN 1x)  & 37.86 & 38.20 & 38.08  \\
Ours (ResNet50-FPN 1x ) & \textbf{38.06}  & \textbf{38.39} & \textbf{38.35} \\
\hline
\end{tabular}
}
\end{center}
\caption{Results on COCO object detection benchmark with our hierarchical cross-entropy.}
\label{tab:faster-rcnn}
\end{table}

With our formulation of a multi-level hierarchical focal loss as described in section~\ref{section:method}, we train a RetinaNet \cite{retinanet} object detector and compare to the baseline focal loss. In Table \ref{tab:coco} we show results on the COCO \cite{mscoco} benchmark, enriched with hierarchies from the WordNet \cite{wordnet} database, pruning the semantic graph to a hierarchical tree similar to \cite{YOLO9000}. 

To evaluate hierarchically, we compute mAP at every level of the hierarchy, where predictions are filtered with non-maximum suppression at each level. An increase in higher-level mAP indicates a decrease in mistake severity of the predictions. As we go up the hierarchy tree, we see that the mAP naturally increases with the reduction of classes, making these predictions suitable for harvesting higher accuracy levels with higher confidence due to aggregation. 

We observe that adding only a few levels of hierarchy to the training yields improvements that propagate to higher levels too. For instance, in Table \ref{tab:coco} we only train with 3 levels of hierarchy, but observe improvements up to level 8. In addition, we find the level weights to be crucial in determining where the biggest improvements will be, in the leaf level or the higher levels. To demonstrate that with our approach it is possible to harvest leaf-level improvements and not only make better mistakes, in Table \ref{tab:coco} we show our results with leaf-focused weighting, namely of 0.8 for the leaf level focal loss, and 0.1 for the next two levels. 

\begin{figure}
\begin{center}
\includegraphics[width=\linewidth]{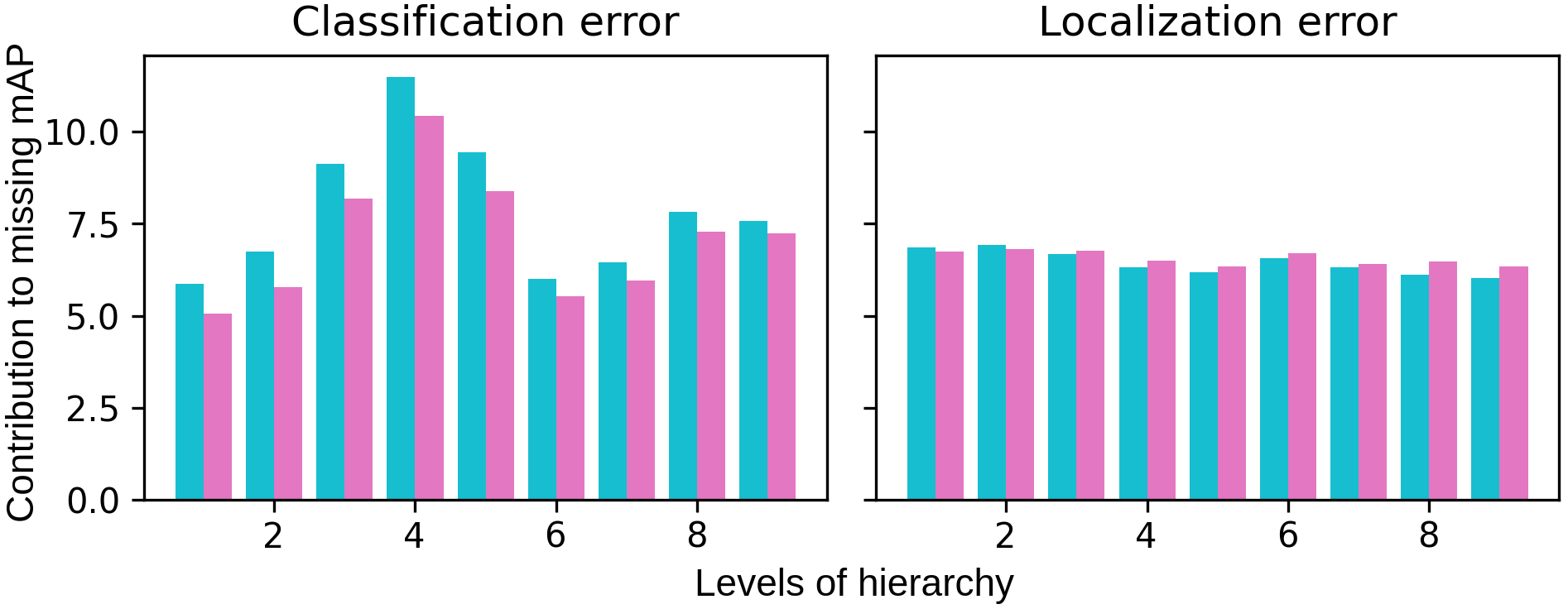}
\caption{Multi-level evaluation effects of the hierarchical loss (purple) compared to the baseline (blue) on classification and localization error computed with TIDE ~\cite{tide-eccv2020}.}
\label{fig:tide_charts}
\end{center}
\end{figure}

In Figure \ref{fig:tide_charts} we plot classification and localization error computed with the TIDE \cite{tide-eccv2020} analysis. TIDE measures the contribution of each error component in the missing mAP. As expected, the hierarchical loss decreases classification error at each hierarchy level, indicating that improvements indeed come from fixing classifier mistakes, while localization performance remains mostly unaffected.

\subsection{Hierarchical cross-entropy for object detection}
To sanity-check the validity of our theoretical framework and the hierarchical cross-entropy loss on softmax-based detection architectures, we experiment with softmax-based Faster-RCNN \cite{ren2015faster}. We train the model with three levels of hierarchies from the WordNet tree. We weigh the softmax cross entropy loss from different levels as \cite{making_better_mistakes} with a weight $ w_l = exp(-\alpha (l-1))$, where $l$ is the level and $\alpha > 0$ is a hyper-parameter that controls the extent to which information is discounted down the hierarchy. We use $\alpha = 0.5$ in the Faster-RCNN hierarchical experiment. As shown in Table \ref{tab:faster-rcnn}, we get improved mAP on all three hierarchical levels. 


\subsection{Hierarchical cross-entropy for classification}



To validate the plausibility of our theory for softmax-based classification, we run experiments on iNaturalist with our hierarchical cross entropy loss. We compare with \cite{making_better_mistakes} in Table \ref{tab:inaturalist} and evaluate on Top-1 error and hierarchical evaluation measures proposed by them.  

We report results from two settings, one with leaf-focused weighting (0.7 for the leaf-level loss and another 0.3 distributed equally among the remaining levels), and a second one with hierarchy-focused weighting (leaf-level weight 0.1 and another 0.9 distributed equally among remaining levels). From the results we observe that with leaf-focused weighting we harvest improvements over cross-entropy on Top-1 accuracy without neglecting the hierarchical metrics, while with hierarchy-focused weighting manage to preserve large chunks of accuracy while focusing on improving hierarchical mistakes. Our conjecture is that improvements over \cite{making_better_mistakes} happen due to our proposed hierarchical-cross entropy being simpler, decoupling the hierarchy levels in the loss computation and leading to cleaner signal from the gradients.